\documentclass[10pt,twocolumn,letterpaper]{article}

\usepackage[letterpaper,margin=19mm,top=21mm,bottom=22mm]{geometry}
\usepackage{mathptmx}
\usepackage{amsmath,amssymb,mathtools}
\usepackage{booktabs}
\usepackage{tabularx}
\usepackage{graphicx}
\usepackage[table]{xcolor}
\usepackage{microtype}
\usepackage[font=small,labelfont=bf]{caption}
\usepackage[authoryear,round]{natbib}
\usepackage[hidelinks]{hyperref}
\usepackage{fancyhdr}
\usepackage{float}

\definecolor{primary}{RGB}{0,8,128}
\definecolor{accent}{RGB}{204,102,0}
\definecolor{softbg}{RGB}{245,248,251}
\hypersetup{
  colorlinks=true,
  linkcolor=primary,
  citecolor=primary,
  urlcolor=primary
}


\newcommand{\norm}[1]{\operatorname{norm}\!\left(#1\right)}
\newcommand{\rel}{\mathrm{rel}}

\newcommand{\ms}[2]{#1\rlap{\tiny\color{black!60}\,$\pm$\,#2}}
\newcommand{\msb}[2]{\textbf{#1}\rlap{\tiny\color{black!60}\,$\pm$\,#2}}

\title{\vspace{-8mm}\rule{\linewidth}{1.6pt}\\[8pt]\Large\bfseries Beyond Isolated Patch Matching: Dual-Scale Common-Mode Suppression for Training-Free Few-Shot Anomaly Detection\\[4pt]\rule{\linewidth}{0.7pt}}

\author{
  \textbf{Ning Zhu}\\
  \small Xi'an University of Technology
}
\date{}

\begin{document}
\maketitle
\thispagestyle{plain}

\begin{center}
  \textbf{\large Abstract}
\end{center}
\vspace{-4pt}
\noindent{\small Nearest-neighbor patch retrieval over frozen DINOv2 features offers a strong training-free baseline for few-shot anomaly detection. Yet isolated memory-bank cosine matching suffers from two forms of \emph{common-mode interference} on the unit hypersphere. First, within each image, self-attention smoothing injects a shared neighborhood average, or spatial common mode, into intermediate tokens. When a defective component remains locally normal in appearance yet violates pose or layout, isolated retrieval still finds a normal match and misses the defect. Second, across the reference bank, strongly anisotropic normal variance, or distributional common mode, inflates baseline cosine similarity and compresses the angular margin for subtle defects, whereas full whitening amplifies tail noise under scarce exemplars. We unify both remedies as \emph{dual-scale common-mode suppression} of the form $I-\alpha P$ followed by unit-sphere re-normalization in a single frozen DINOv2 forward pass. At the micro scale, hard spatial removal ($\alpha{=}1$, $P{=}c_i c_i^\top$) strips each mid-layer token's attended context to expose patch-to-context structural contrast, scored via reference-only calibration over only the top-$5$ appearance matches. At the macro scale, soft distributional attenuation ($0<\alpha<1$, $P{=}UU^\top$) dampens dominant reference variance offline while preserving mean orientation and coarse part alignment. Across MVTec-AD and VisA from $1$-shot to $16$-shot, our method consistently outperforms prior training-free baselines such as AnomalyDINO and SubspaceAD-R, achieving up to $98.9\%$ I-AUROC on MVTec-AD and $+3.6$ pixel $F_1$-max on VisA at under $8\%$ runtime overhead.
}\vspace{4pt}

\section{Introduction}
\label{sec:intro}

Industrial visual anomaly detection identifies and localizes surface or structural defects from only a handful of normal exemplars \citep{bergmann2019mvtec,zou2022visa}. Recent training-free detectors built on frozen self-supervised Vision Transformers \citep{dosovitskiy2021vit} such as DINOv2 \citep{oquab2024dinov2,damm2025anomalydino} show that nearest-neighbor patch retrieval on the unit hypersphere rivals complex multimodal pipelines without optimizing a single weight \citep{caron2021dino,zhou2022ibot}.

\begin{figure*}[!t]
  \centering
  \includegraphics[width=0.84\textwidth]{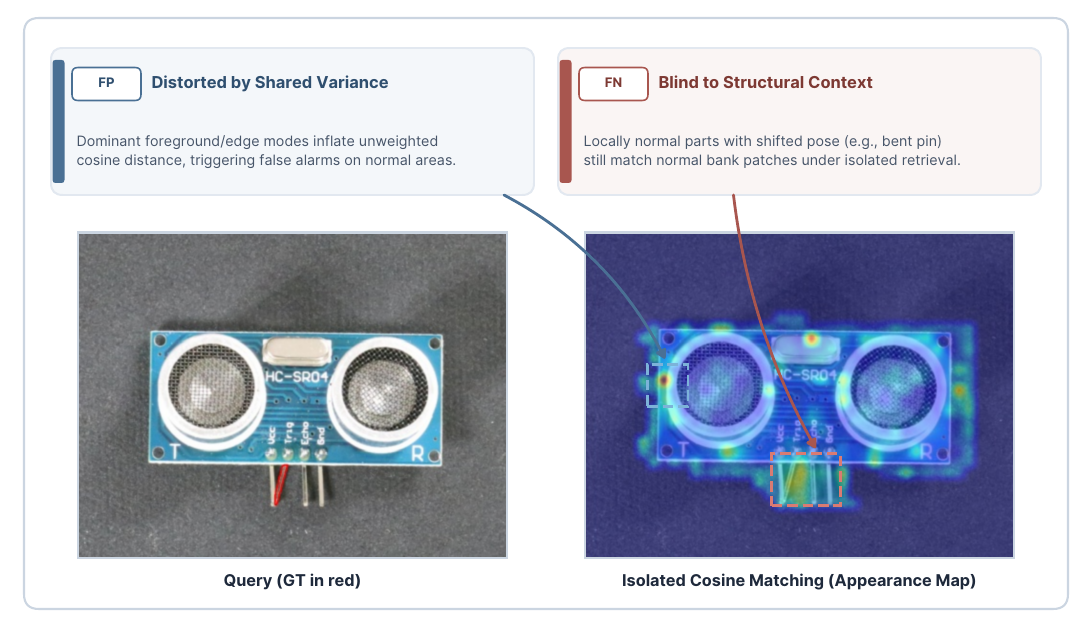}
  \caption{Isolated few-shot cosine matching suffers simultaneously from false positives and false negatives. On a real VisA \emph{pcb1} query (bottom left, ground-truth bent pin in red), dominant shared variance inflates unweighted cosine distance on normal edges (blue dashed box), while the bent pin matches locally normal patches and is missed (red dashed box).}
  \label{fig:motivation}
\end{figure*}

Yet standard isolated cosine matching over frozen DINOv2 patches suffers from two forms of \emph{common-mode interference} that compress discriminative angular margins (Figure~\ref{fig:motivation}). First, within each image, spatial common mode blinds isolated matching to structural context. On a circuit board with a bent connector pin or a cable with a swapped wire, the defective part stays locally normal in appearance and still retrieves a normal bank patch. In raw intermediate ViT tokens, self-attention smoothing further mixes nearby patches into a shared neighborhood average \citep{darcet2024registers}, which hides the relative layout contrast that signals a shifted pose. Second, across the reference bank, distributional common mode distorts unweighted cosine distance. DINOv2 patch embeddings are strongly anisotropic \citep{ethayarajh2019anisotropy,oquab2024dinov2}, with leading principal directions dominated by foreground--background separation and coarse part variation shared across normal and defective regions alike. Under unweighted cosine similarity, these high-variance shared directions inflate baseline similarity and narrow the margin for subtle surface defects \citep{mu2018allbutthetop}. Full covariance whitening is not a viable fix under few-shot scarcity because it overfits ill-conditioned tail directions.

\begin{figure*}[!t]
  \centering
  \includegraphics[width=0.99\textwidth]{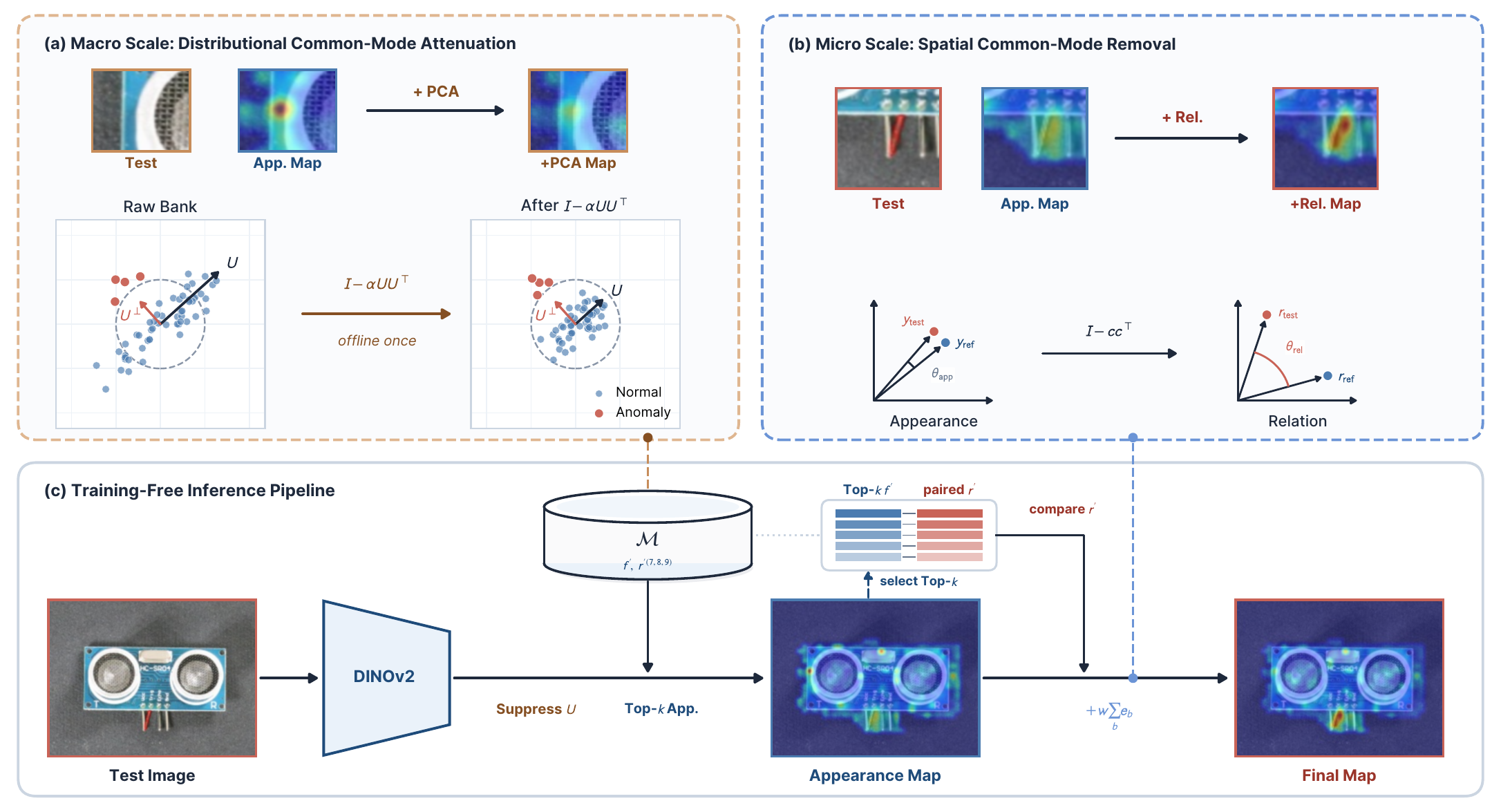}
  \caption{Dual-scale common-mode suppression unifies bank-level and token-level geometric conditioning in one frozen pass. (a) Macro scale ($0<\alpha<1$): offline reference PCA attenuation ($I-\alpha UU^\top$) compresses shared variance $U$ while preserving orthogonal defect directions $U^\perp$ and mean $\mu$. (b) Micro scale ($\alpha=1$): hard spatial removal ($I-c c^\top$) strips neighborhood smoothing to expose shifted layouts ($\theta_{\mathrm{rel}} \gg \theta_{\mathrm{app}}$). (c) Inference applies frozen $(\mu, U)$, retrieves top-$5$ appearance matches from the joint bank, and adds calibrated relational residuals.}
  \label{fig:mechanism}
\end{figure*}

We resolve both failure modes without training or abandoning memory-bank retrieval \citep{lendering2026subspacead}. Both remedies follow one geometric operator: \emph{dual-scale common-mode suppression} of the form $I-\alpha P$ followed by unit-sphere re-normalization in a single frozen DINOv2 forward pass (Figure~\ref{fig:mechanism}). Rather than reducing dimensionality, both operators dampen dominant shared components so that re-normalization onto $\mathbb{S}^{D-1}$ expands the angular separation of true anomalies. At the micro (token) scale, we hook mid-layer self-attention to form each patch's attended context $c_i$ and apply hard spatial common-mode removal ($\alpha{=}1$, $P{=}c_i c_i^\top$) before L2 re-normalization. Stripping this shared neighborhood average amplifies patch-to-context structural contrast, which we score via reference-only quantile calibration over the top-$5$ appearance matches which adds $O(1)$ extra scoring complexity per patch and $O(N)$ per image, independent of memory-bank size. At the macro (bank) scale, we estimate dominant normal variance directions $U$ once offline from the reference bank and apply soft distributional attenuation ($0<\alpha<1$, $P{=}UU^\top$) around the bank mean $\mu$ to both appearance and relational streams. Soft attenuation compresses loud shared variation while preserving full dimensionality and coarse part-alignment anchors, which avoids both the tail-noise amplification of whitening and the semantic erasure of hard removal ($\alpha{=}1$) under strictly inductive inference \citep{li2024musc}.

With only $O(N)$ per-image ($O(1)$ per-patch) extra scoring overhead, adding $7.7\%$--$8.1\%$ latency, dual-scale common-mode suppression outperforms prior memory-bank, vision-language, and subspace-modeling detectors across $1$-shot to $16$-shot regimes on MVTec-AD and VisA \citep{jeong2023winclip,damm2025anomalydino,lendering2026subspacead}. Our method reaches up to $98.9\%$ I-AUROC, $97.8\%$ P-AUROC, and $94.9\%$ AU-PRO on MVTec-AD, while lifting VisA by $+1.0$ I-AUROC, $+3.6$ pixel $F_1$-max, and $+1.3$ AU-PRO on average over the same-backbone baseline. Suppressing spatial and distributional common modes thus unlocks substantial accuracy in frozen memory-bank retrieval at minimal runtime cost.

\section{Related Work}
\label{sec:related}

\paragraph{Reconstruction and Distillation Methods.}
Reconstruction-based detectors score anomalies via image- or feature-level reconstruction residuals, including synthetic-defect frameworks (CutPaste \citep{li2021cutpaste}, DR{\AE}M \citep{zavrtanik2021draem}, SimpleNet \citep{liu2023simplenet}, RealNet \citep{zhang2024realnet}), multi-class transformers and diffusion models (UniAD \citep{you2022uniad}, HVQ-Trans \citep{lu2023hvqtrans}, GLAD \citep{yao2024glad}, VPDM \citep{li2024vpdm}), and feature-distillation or pure-Transformer reconstructors (RD4AD \citep{deng2022rd4ad}, RD++ \citep{tien2023rdpp}, DeSTSeg \citep{zhang2023destseg}, Dinomaly \citep{guo2025dinomaly}, INP-Former \citep{luo2025inpformer}). While effective in full-shot regimes, these methods require task-specific training and face a fidelity--stability trade-off under scarce exemplars \citep{zhang2026rad}.

\paragraph{Memory-Bank, Statistical, and Foundation-Model Methods.}
Memory-bank detectors bypass training by matching test patches against normal descriptors, from PatchCore \citep{roth2022patchcore} and its extensions (SoftPatch \citep{jiang2022softpatch}, CFA \citep{lee2022cfa}, ReConPatch \citep{hyun2024reconpatch}, PNI \citep{bae2023pni}) to few-shot methods such as RegAD \citep{huang2022regad}, GraphCore \citep{xie2023graphcore}, and CostFilter-AD \citep{zhang2025costfilter}. Analytical geometry approaches model normal distributions via hyperspheres (Deep SVDD \citep{ruff2018deepsvdd}), per-patch Gaussians (PaDiM \citep{defard2021padim}), normalizing flows (FastFlow \citep{yu2021fastflow}, CFLOW-AD \citep{gudovskiy2022cflow}), ridge regression (FastRecon \citep{fang2023fastrecon}), or PCA subspaces (SubspaceAD \citep{lendering2026subspacead}). Foundation-model detectors leverage CLIP \citep{radford2021clip}, SAM \citep{kirillov2023sam}, and DINOv2 \citep{oquab2024dinov2}, including vision-language prompt methods (WinCLIP \citep{jeong2023winclip}, AnomalyCLIP \citep{zhou2024anomalyclip}, PromptAD \citep{li2024promptad}, AnomalyGPT \citep{gu2024anomalygpt}), transductive zero-shot scoring (MuSc \citep{li2024musc}), and trained adapters (FOUNDAD \citep{zhai2026foundad}). Most closely related to our work, AnomalyDINO \citep{damm2025anomalydino} and RAD \citep{zhang2026rad} perform training-free cosine retrieval over frozen DINOv2 patch banks \citep{darcet2024registers}, while \citet{mu2018allbutthetop} and \citet{ethayarajh2019anisotropy} highlight Transformer anisotropy. Unlike SubspaceAD, which replaces patch retrieval with subspace residuals, or hard removal at $\alpha{=}1$, we retain inductive memory-bank retrieval and unify micro-scale spatial common-mode removal with macro-scale soft attenuation ($0<\alpha<1$) in a single forward pass.

\section{Preliminaries}
\label{sec:prelim}

\paragraph{Task and Frozen Patch Descriptors.}
Given $k \in \{1, 2, 4, 8, 16\}$ normal reference images of a category and no anomalous samples, training-free few-shot anomaly detection assigns a pixel-level anomaly map and image-level score to each test image without fine-tuning. We extract dense patch descriptors from a frozen DINOv2 ViT-B/14 backbone \citep{oquab2024dinov2} ($D=768$). Images are resized to $448$\,px on the shorter edge and cropped to a multiple of $14$, yielding an $H_p \times W_p$ grid of $N = H_p W_p$ tokens, with $N=1024$ on $448{\times}448$ inputs. The \emph{appearance descriptor} of patch $i$ is the L2-normalized last-block post-LayerNorm token:
\begin{equation}
  f_i = \frac{\phi_{\text{last}}(x)_i}{\lVert \phi_{\text{last}}(x)_i \rVert_2} \in \mathbb{S}^{D-1},
\end{equation}
forming $F \in \mathbb{R}^{N \times D}$ in one forward pass. Following \citet{damm2025anomalydino}, each reference image is augmented via a per-category rotation/flip policy into $V$ views and reference bank $\mathcal{F} = \{f_{v,p}\}$ of $V \cdot N$ descriptors, while test images are evaluated unrotated with a $1$-component PCA foreground mask $M$ refined morphologically. Reference images remain unmasked, reinforcing the need for macro-scale suppression (Section~\ref{sec:macro}).

\paragraph{Appearance Nearest-Neighbor Baseline.}
For a foreground test patch $q \in M$, the appearance score is half the squared Euclidean distance on $\mathbb{S}^{D-1}$ to its nearest reference patch, while background patches with $q \notin M$ receive $a(q)=0$:
\begin{equation}
  a(q) = 1 - \max_{p \in \mathcal{F}} \cos(q, p) = \min_{p \in \mathcal{F}} \frac{1}{2}\lVert q - p \rVert_2^2.
\end{equation}
Following AnomalyDINO, the image-level score averages the top $\lfloor 0.01 N \rfloor$ patch scores:
\begin{equation}
  S_{\text{img}} = T_{1\%}\big(\{a(q)\}_{q=1}^{N}\big) = \frac{1}{\lfloor 0.01 N \rfloor} \sum_{x \in \text{top-}\lfloor 0.01 N \rfloor} x.
\end{equation}

\section{Method: Dual-Scale Common-Mode Suppression}
\label{sec:method}

Our framework strengthens nearest-neighbor memory-bank retrieval with two modular linear common-mode suppression operators of the form $I - \alpha P$ followed by unit-sphere re-normalization, where $I \in \mathbb{R}^{D\times D}$ denotes the identity matrix: (1) a \emph{micro-scale} hard spatial common-mode removal ($\alpha = 1$, $P = c_i c_i^\top$) that strips each mid-layer token's attended neighborhood average to expose patch-to-context contrast (Section~\ref{sec:micro}); and (2) a \emph{macro-scale} soft distributional common-mode attenuation ($\alpha \in (0,1)$, $P = U U^\top$) that dampens high-variance reference directions while preserving full feature dimensionality (Section~\ref{sec:macro}), coupled via leakage-free reference calibration (Section~\ref{sec:calib}). Because both operators attenuate shared background components before L2 re-normalization, they expand the spherical angular margin for anomalies without training. Figure~\ref{fig:mechanism}(c) illustrates the full inference pipeline.

\subsection{Micro Scale: Spatial Common-Mode \mbox{Removal}}
\label{sec:micro}

\paragraph{Attention Context and Rank-$1$ Orthogonal Projection.}
Raw intermediate ViT patch tokens remain strongly coupled to local appearance and shared neighborhood semantics, heavily overlapping with the final-layer appearance stream $f_i$ while overshadowing the fine-grained relative difference between patch $i$ and its surroundings. To isolate and amplify this relative signal without training, we attach two lightweight forward hooks to three consecutive late-intermediate blocks $\mathcal{B} = \{7, 8, 9\}$, where semantic structure is salient while spatial resolution remains intact.

For block $b \in \mathcal{B}$, let $Q, K$ denote the frozen query and key projections computed from the attention input. We obtain the head-averaged self-attention matrix restricted to the $N$ patch tokens, retaining self-attention self-weights, and row-normalize each row to sum to $1$:
{\small\begin{equation}
  A^{(b)} = \operatorname{row\_norm}\!\left(\operatorname{mean}_{\text{heads}}\!\operatorname{softmax}\!\left(\frac{Q K^\top}{\sqrt{d}}\right)\right) \in \mathbb{R}^{N \times N}.
\end{equation}}
Let $y_i^{(b)} \in \mathbb{S}^{D-1}$ be the L2-normalized post-LayerNorm output of block $b$ at patch $i$. The \emph{attention context} $c_i^{(b)} \in \mathbb{S}^{D-1}$ is the unit-normalized attention-weighted mean of the patch features it attends to:
\begin{equation}
  c_i^{(b)} = \norm{\sum_{j=1}^{N} A_{ij}^{(b)}\, y_j^{(b)}}.
\end{equation}
Because $c_i^{(b)}$ has unit norm, the outer product $P_{\text{loc}} = c_i^{(b)} c_i^{(b)\top} \in \mathbb{R}^{D\times D}$ is the exact rank-$1$ orthogonal projector onto the context direction. Applying the complementary linear operator $(I - c_i^{(b)} c_i^{(b)\top})$ to $y_i^{(b)}$ and re-normalizing onto $\mathbb{S}^{D-1}$ yields the \emph{relation descriptor}:
\begin{align}
  r_i^{(b)} &= \norm{\big(I - c_i^{(b)} c_i^{(b)\top}\big)\, y_i^{(b)}} \nonumber \\
            &= \norm{y_i^{(b)} - \big(y_i^{(b)\top} c_i^{(b)}\big)\, c_i^{(b)}}.
\end{align}
Geometrically, $(I - c_i^{(b)} c_i^{(b)\top})$ zeros out the component of $y_i^{(b)}$ collinear with its attended context $c_i^{(b)}$, and the subsequent L2 re-normalization expands the surviving orthogonal residual to full unit-sphere magnitude. What emerges is an amplified \emph{patch-to-context relational contrast}, capturing relative layout, boundary continuity, and local structural composition, extracted in the exact same single forward pass as $f_i$.

\paragraph{Matched Relation Distance.}
We do \emph{not} run an independent nearest-neighbor search in relation space. Instead, we reuse the query's top-$K_{\text{nn}}$ ($K_{\text{nn}} = 5$) appearance matches $\mathcal{M}(q) = \{p_1, \dots, p_5\} \subset \mathcal{F}'$ and compute:
\begin{align}
  \rel_b(q) &= \min_{p \in \mathcal{M}(q)} \Big(1 - \cos\big(r_b(q), r_b(p)\big)\Big) \nonumber \\
            &= \min_{p \in \mathcal{M}(q)} \frac{1}{2}\big\lVert r_b(q) - r_b(p)\big\rVert_2^2.
\end{align}
Reusing $\mathcal{M}(q)$ verifies whether a test patch exhibits normal relational contrast relative to \emph{the same normal counterparts} that explain its appearance. Taking the minimum over $5$ candidates accepts any semantically plausible counterpart (whereas averaging over $\mathcal{M}(q)$ would dilute the anomaly signal).

\subsection{Macro Scale: Distributional Common-Mode Attenuation}
\label{sec:macro}

\paragraph{Few-Shot Anisotropy and Soft Attenuation.}
Across a $k$-shot reference bank ($k \in \{1,2,4,8,16\}$), descriptors are scarce and strongly anisotropic. Reflecting DINOv2's pretraining geometry \citep{oquab2024dinov2}, the leading principal directions of an unmasked category bank concentrate on foreground--background separation, global illumination/texture, and coarse part-level semantics shared across normal and defective regions alike. In cosine geometry, large shared projections along these high-variance axes inflate baseline similarity and compress fine-grained defect contrast \citep{mu2018allbutthetop}.

For each descriptor space $X = \{x\}$ covering appearance $f$ and each relation block $r^{(b)}$ over all $V \cdot N$ reference patches of a category, we fit a dedicated PCA on the centered reference covariance:
\begin{equation}
  \mu = \frac{1}{|X|}\sum_{x \in X} x, \qquad
  \Sigma = \frac{1}{|X|}\sum_{x \in X} (x - \mu)(x - \mu)^\top,
\end{equation}
and collect the top-$k_{\text{pca}}$ orthonormal eigenvectors $U = [u_1, \dots, u_{k_{\text{pca}}}] \in \mathbb{R}^{D \times k_{\text{pca}}}$ ($U^\top U = I_{k_{\text{pca}}}$). Here $P_{\text{glob}} = U U^\top \in \mathbb{R}^{D\times D}$ is the orthogonal projector onto the $k_{\text{pca}}$-dimensional high-variance reference subspace. We apply a \emph{soft subspace suppression} operator with strength $\alpha \in (0, 1]$ followed by L2 re-normalization onto $\mathbb{S}^{D-1}$:
\begin{align}
  x' &= \norm{\mu + \big(I - \alpha\, U U^\top\big)(x - \mu)} \nonumber \\
     &= \norm{x - \alpha\, U U^\top (x - \mu)} \in \mathbb{S}^{D-1}.
\end{align}
Decomposing the centered vector $\tilde{x} = x - \mu = \tilde{x}_U + \tilde{x}_{U^\perp}$ makes the metric conditioning effect transparent: $(I - \alpha U U^\top)\tilde{x} = (1-\alpha)\tilde{x}_U + \tilde{x}_{U^\perp}$. Components along the loudest shared reference directions are attenuated by $(1-\alpha)$, while all low-variance orthogonal directions remain untouched. Retaining $\mu$ prior to L2 re-normalization anchors transformed descriptors in the category's original directional hemisphere to preserve coarse part orientation, unlike mean-centering postprocessing \citep{mu2018allbutthetop}. Unlike full whitening that amplifies noisy tail eigenvalues or hard removal at $\alpha = 1$ that erases useful cues, partial suppression with $k_{\text{pca}} = 64$ and $\alpha = 0.5$ improves contrast stably across seeds and categories.

To keep retrieval geometry aligned, we fit separate reference-side PCAs on the appearance stream $f$ and on each relation block $r^{(b)}$ over all augmented reference patches, and apply the resulting frozen $(\mu, U, \alpha)$ identically to the reference bank $\mathcal{F}'$, reference calibration descriptors, and test queries.

\subsection{Score Fusion, Algorithm, and Complexity}
\label{sec:calib}

\paragraph{Reference-Only Score Calibration and Fusion.}

Appearance descriptors $f$ and orthogonal relation residuals $r^{(b)}$ inhabit different subspaces with distinct baseline distances and tail spreads, even normal reference patches exhibit non-zero relation distances that vary by category and block. Directly summing $a(q)$ and $\rel_b(q)$ therefore leaves the two streams uncalibrated.

To align both streams without touching any test data, we estimate per-block calibration parameters strictly from \emph{reference--reference} pairs in $\mathcal{F}'$:
\begin{enumerate}
  \item For every reference patch $p \in \mathcal{F}'$, retrieve its $2$nd--$6$th nearest neighbors in appearance space excluding the self-match, recording its $2$-NN squared Euclidean distance $d^{(2)}(p)$ on $\mathbb{S}^{D-1}$, where $d^{(2)}(p)/2 = 1 - \cos(p, p_{(2)})$, and matched relation distance $\rel_b(p)$ over those $5$ neighbors.
  \item Compute the empirical $95$th and $99$th percentiles of the normal reference relation distances and the reference appearance scale:
\begin{align}
  \tau_b &= Q_{0.95}\big(\{\rel_b(p)\}\big), \label{eq:tau} \\
  \sigma_b &= \max\big(Q_{0.99}(\{\rel_b(p)\}) - \tau_b,\, 0.01\big), \label{eq:sigma} \\
  s_{\text{app}} &= \max\big(Q_{0.95}(d^{(2)}/2),\, 10^{-6}\big). \label{eq:sapp}
\end{align}
\end{enumerate}
At test time, only relational distances exceeding $\tau_b$ contribute, normalized by the tail spread $\sigma_b$ and rescaled into appearance units via $s_{\text{app}}$:
\begin{equation}
  e_b(q) = \frac{\max\big(0,\, \rel_b(q) - \tau_b\big)}{\sigma_b} \cdot s_{\text{app}}.
\end{equation}
The final patch anomaly score sums the calibrated excess over blocks $\mathcal{B} = \{7, 8, 9\}$:
\begin{equation}
  s(q) = a(q) + w \sum_{b \in \{7,8,9\}} e_b(q), \qquad w = 0.25.
\end{equation}
Summing rather than averaging across $\mathcal{B}$ allows strong relational evidence at any single block to elevate the score without per-layer weight tuning. For pixel-level evaluation, patch scores with background zeroed are reshaped to $H_p \times W_p$, bilinearly upsampled to $448\times 448$, and smoothed with a Gaussian kernel ($\sigma = 4$).

\paragraph{Algorithm and Complexity.}
Algorithm~1 summarizes reference preprocessing and inductive test-time inference.

\begin{table}[!htbp]
  \centering
  \begin{minipage}{\linewidth}
    \small
    \hrule height 0.9pt
    \vspace{3pt}
    \noindent\textbf{Algorithm 1} Dual-Scale Common-Mode Anomaly Scoring
    \vspace{3pt}
    \hrule height 0.5pt
    \vspace{4pt}
    \par\noindent\textbf{Input:} $k$ reference images $\{x_m^{\text{ref}}\}$, test image $x^{\text{test}}$, frozen ViT $\phi$, blocks $\mathcal{B}=\{7,8,9\}$, $k_{\text{pca}}=64, \alpha=0.5, w=0.25$.
    \par\vspace{1pt}\noindent\textbf{Output:} Image score $S_{\text{img}}$, pixel anomaly map $\hat{S}_{\text{pix}}$.
    \par\vspace{4pt}\noindent\textit{// Stage 1: Reference Preprocessing (once per category)}
    \begin{list}{}{\leftmargin=1.6em \itemindent=-1.6em \topsep=2pt \partopsep=0pt \parsep=0pt \itemsep=1.5pt}
      \item 1:\quad Augment $\{x_m^{\text{ref}}\}$ via per-category rotation/flip policy $\to V$ views.
      \item 2:\quad \textbf{for} each view $v = 1,\dots,V$ \textbf{do}
      \item \hspace{1.2em}3:\quad Extract $f_v$ and $\{y_v^{(b)}, A_v^{(b)}\}_{b\in\mathcal{B}}$ in one forward pass.
      \item \hspace{1.2em}4:\quad Compute $c_{v,i}^{(b)}$ and $r_{v,i}^{(b)} \leftarrow \norm{(I - c_{v,i}^{(b)}c_{v,i}^{(b)\top})y_{v,i}^{(b)}}$.
      \item 5:\quad \textbf{end for}
      \item 6:\quad Fit PCA $(\mu_f, U_f)$ on $\{f\}$ and $(\mu_b, U_b)$ on $\{r^{(b)}\}_{b\in\mathcal{B}}$.
      \item 7:\quad Apply $x' \leftarrow \norm{x - \alpha U U^\top (x-\mu)}$ to all references $\to \mathcal{F}'$.
      \item 8:\quad \textbf{for} each reference patch $p \in \mathcal{F}'$ \textbf{do}
      \item \hspace{1.2em}9:\quad Find $2$nd--$6$th NNs in $\mathcal{F}'$; compute $d^{(2)}(p)$ and $\rel_b(p)$.
      \item 10:\quad \textbf{end for}
      \item 11:\quad Compute $s_{\text{app}}, \{\tau_b, \sigma_b\}_{b\in\mathcal{B}}$ via Eqs.~\eqref{eq:tau}--\eqref{eq:sapp}.
    \end{list}
    \par\vspace{3pt}\noindent\textit{// Stage 2: Inductive Test Inference (per test image)}
    \begin{list}{}{\leftmargin=1.6em \itemindent=-1.6em \topsep=2pt \partopsep=0pt \parsep=0pt \itemsep=1.5pt}
      \item 12:\quad Forward $x^{\text{test}}$ once; extract $f, \{r^{(b)}\}_{b\in\mathcal{B}}$ and PC1 mask $M$.
      \item 13:\quad Apply reference $(\mu, U, \alpha)$ to test $f$ and $\{r^{(b)}\}_{b\in\mathcal{B}}$.
      \item 14:\quad \textbf{for} each foreground patch $q \in M$ \textbf{do}
      \item \hspace{1.2em}15:\quad $a(q) \leftarrow 1 - \max_{p\in\mathcal{F}'} \cos(q, p)$; store top-$5$ matches $\mathcal{M}(q)$.
      \item \hspace{1.2em}16:\quad \textbf{for} $b \in \mathcal{B}$ \textbf{do}
      \item \hspace{2.4em}17:\quad $\rel_b(q) \leftarrow \min_{p\in\mathcal{M}(q)}(1 - \cos(r_b(q), r_b(p)))$.
      \item \hspace{2.4em}18:\quad $e_b(q) \leftarrow \max(0, \rel_b(q) - \tau_b)/\sigma_b \cdot s_{\text{app}}$.
      \item \hspace{1.2em}19:\quad \textbf{end for}
      \item \hspace{1.2em}20:\quad $s(q) \leftarrow a(q) + w \sum_{b\in\mathcal{B}} e_b(q)$.
      \item 21:\quad \textbf{end for}
      \item 22:\quad $S_{\text{img}} \leftarrow T_{1\%}(\{s(q)\}_{q=1}^{N})$; upsample \& smooth $s \to \hat{S}_{\text{pix}}$.
    \end{list}
    \vspace{3pt}
    \hrule height 0.9pt
  \end{minipage}
\end{table}

Per-image inference requires one ViT forward pass and one dense nearest-neighbor lookup in appearance space of cost $O(N \cdot |\mathcal{F}'| \cdot D)$. Because relation scoring evaluates only the fixed $K_{\text{nn}}{=}5$ appearance matches in $\mathcal{M}(q)$, both operators add only $O(1)$ extra complexity per patch, or $O(N)$ per image, independent of memory-bank size $|\mathcal{F}'|$. Reference-side PCA eigendecomposition costs $O(D^3)$ once per category and is amortized across all test images.

\section{Experiments}
\label{sec:experiments}

\subsection{Experimental Setup}
\label{sec:exp_setup}

\paragraph{Benchmarks and Protocol.}
We evaluate on \textbf{MVTec-AD} \citep{bergmann2019mvtec} ($15$ categories) and \textbf{VisA} \citep{zou2022visa} ($12$ categories, official $1$-class split) across $k \in \{1, 2, 4, 8, 16\}$ shots using the frozen DINOv2 ViT-B/14 pipeline from Section~\ref{sec:prelim}. Pixel maps are smoothed with a Gaussian kernel ($\sigma{=}4$) and evaluated on a $448{\times}448$ canvas for MVTec-AD with nearest-neighbor resized masks, and at native resolution for VisA following \citet{damm2025anomalydino}. Reference sets follow a strictly nested $3$-seed protocol ($s \in \{42, 43, 44\}$, $1 \subset 2 \subset 4 \subset 8 \subset 16$); we compute per-category means first and report $3$-seed mean $\pm$ std (in $\%$) across image-level (\textbf{I-AUROC}, \textbf{I-$F_1$-max}, \textbf{I-AP}) and pixel-level (\textbf{P-AUROC}, \textbf{P-$F_1$-max}, \textbf{AU-PRO} at $\text{FPR} \le 0.3$) metrics.

\paragraph{Hyperparameters.}
A single global configuration is fixed across all categories, shot counts, and benchmarks: $\mathcal{B}=\{7,8,9\}$, $K_{\text{nn}}=5$, $w=0.25$, $k_{\text{pca}}=64$, $\alpha=0.5$, and calibration quantiles $(0.95, 0.99, 0.95)$.

\subsection{Main Results on MVTec-AD and VisA}
\label{sec:exp_main}

We compare against representative few-shot detectors under the protocol of \citet{damm2025anomalydino}: SPADE \citep{cohen2020spade}, PaDiM \citep{defard2021padim}, PatchCore \citep{roth2022patchcore}, FastRecon \citep{fang2023fastrecon}, WinCLIP+ \citep{jeong2023winclip}, APRIL-GAN \citep{chen2023april}, ADP$_\ell$ \citep{kwak2024adp}, AnomalyDINO \citep{damm2025anomalydino}, and SubspaceAD \citep{lendering2026subspacead}. For AnomalyDINO$^*$ \citep{damm2025anomalydino}, Table~\ref{tab:main_results} reports our unified DINOv2 ViT-B/14 reproduction at $448$\,px over $3$ seeds, matching the Appearance baseline in Table~\ref{tab:comp_ablation}. For SubspaceAD \citep{lendering2026subspacead}, we include our unified $3$-seed nested $448$\,px reproduction SubspaceAD-R$^*$ using its official default \texttt{dinov2-with-registers-base} backbone across $1$--$16$-shot. Results for all other published methods are taken directly from \citet{damm2025anomalydino} and their original publications ($^\dagger$ from \citet{jeong2023winclip}; $^\S$ from \citet{kwak2024adp}).

\begin{table*}[!t]
  \centering
  \caption{Few-shot anomaly detection and localization results on \textbf{MVTec-AD} and \textbf{VisA} (in $\%$). $^\dagger$From \citet{jeong2023winclip}; $^\ddagger$cross-dataset supervised linear layers; $^\S$from \citet{kwak2024adp}; $^*$our unified nested $448$\,px reproduction (SubspaceAD-R$^*$ uses the official default \texttt{dinov2-with-registers-base} backbone). Best in \textbf{bold}; second-best \underline{underlined}.}
  \label{tab:main_results}
  \scriptsize
  \renewcommand{\arraystretch}{1.06}
  \setlength{\tabcolsep}{9.5pt}
  \resizebox{\textwidth}{!}{%
  \begin{tabular}{llcccccc@{\hspace{24pt}}cccccc@{\hspace{10pt}}}
    \toprule
    & & \multicolumn{6}{c}{\textbf{MVTec-AD} ($15$ categories)} & \multicolumn{6}{c}{\textbf{VisA} ($12$ categories)} \\
    \cmidrule(lr){3-8} \cmidrule(lr){9-14}
    & & \multicolumn{3}{c}{Image-level} & \multicolumn{3}{c}{Pixel-level} & \multicolumn{3}{c}{Image-level} & \multicolumn{3}{c}{Pixel-level} \\
    \cmidrule(lr){3-5} \cmidrule(lr){6-8} \cmidrule(lr){9-11} \cmidrule(lr){12-14}
    \textbf{Setup} & \textbf{Method} & AUC & $F_1$ & AP & AUC & $F_1$ & PRO & AUC & $F_1$ & AP & AUC & $F_1$ & PRO \\
    \midrule
    1-shot
      & SPADE$^\dagger$ & 81.0 & 90.3 & 90.6 & 91.2 & 42.4 & 83.9 & 79.5 & 80.7 & 82.0 & 95.6 & 35.5 & 84.1 \\
      & PaDiM$^\dagger$ & -- & -- & -- & -- & -- & -- & 62.8 & 75.3 & 68.3 & 89.9 & 17.4 & 64.3 \\
      & PatchCore$^\dagger$ & 83.4 & 90.5 & 92.2 & 92.0 & 50.4 & 79.7 & 79.9 & 81.7 & 82.8 & 95.4 & 38.0 & 80.5 \\
      & FastRecon & 85.7 & -- & -- & 93.2 & -- & -- & 76.2 & -- & -- & 96.7 & -- & -- \\
      & WinCLIP+ & 93.1 & 93.7 & 96.5 & 95.2 & 55.9 & 87.1 & 83.8 & 83.1 & 85.1 & 96.4 & 41.3 & 85.1 \\
      & APRIL-GAN$^\ddagger$ & 92.0 & 92.4 & 95.8 & 95.1 & 54.2 & 90.6 & \underline{91.2} & 86.9 & \underline{93.3} & 96.0 & 38.5 & 90.0 \\
      & SubspaceAD-R$^*$ & 94.6 & \underline{95.9} & \underline{97.4} & \textbf{97.2} & \underline{57.9} & \textbf{93.9} & 89.5 & 86.9 & 88.6 & \underline{97.7} & 38.1 & \textbf{93.4} \\
      & AnomalyDINO$^*$ & \underline{95.4} & 95.5 & 97.0 & 96.0 & 55.7 & 91.9 & 91.0 & \underline{87.5} & 92.5 & 97.3 & \underline{44.5} & 91.7 \\
      & \textbf{Ours} & \msb{96.4}{0.2} & \msb{96.7}{0.3} & \msb{98.3}{0.2} & \msb{97.2}{0.1} & \msb{59.5}{0.2} & \msb{93.9}{0.1} & \msb{93.0}{0.3} & \msb{89.4}{0.1} & \msb{94.2}{0.4} & \msb{97.9}{0.0} & \msb{48.9}{0.3} & \msb{93.4}{0.2} \\
    \midrule
    2-shot
      & SPADE$^\dagger$ & 82.9 & 91.1 & 91.7 & 92.0 & 44.5 & 85.7 & 80.7 & 81.7 & 82.3 & 96.2 & 40.5 & 85.7 \\
      & PaDiM$^\dagger$ & -- & -- & -- & -- & -- & -- & 67.4 & 75.7 & 71.6 & 92.0 & 21.1 & 70.1 \\
      & PatchCore$^\dagger$ & 86.3 & 92.0 & 93.8 & 93.3 & 53.0 & 82.3 & 81.6 & 82.5 & 84.8 & 96.1 & 41.0 & 82.6 \\
      & FastRecon & 88.3 & -- & -- & 94.5 & -- & -- & 86.1 & -- & -- & 97.6 & -- & -- \\
      & WinCLIP+ & 94.4 & 94.4 & 97.0 & 96.0 & 58.4 & 88.4 & 84.6 & 83.0 & 85.8 & 96.8 & 43.5 & 86.2 \\
      & APRIL-GAN$^\ddagger$ & 92.4 & 92.6 & 96.0 & 95.5 & 55.9 & 91.3 & 92.2 & 87.7 & \underline{94.2} & 96.2 & 39.3 & 90.1 \\
      & ADP$_\ell$ & 95.4 & -- & -- & -- & -- & -- & 86.9 & -- & -- & -- & -- & -- \\
      & SubspaceAD-R$^*$ & 96.2 & \underline{96.7} & \underline{98.1} & \textbf{97.4} & \underline{59.6} & \textbf{94.3} & 90.5 & 87.8 & 89.5 & \underline{97.8} & 39.4 & \textbf{93.9} \\
      & AnomalyDINO$^*$ & \underline{96.8} & 96.5 & 98.0 & 96.2 & 57.2 & 92.4 & \underline{92.9} & \underline{89.1} & 94.0 & 97.5 & \underline{46.7} & 92.4 \\
      & \textbf{Ours} & \msb{97.5}{0.8} & \msb{97.2}{0.1} & \msb{98.8}{0.4} & \underline{\ms{97.3}{0.1}} & \msb{60.5}{0.7} & \underline{\ms{94.1}{0.3}} & \msb{94.2}{0.4} & \msb{90.4}{0.4} & \msb{95.1}{0.4} & \msb{98.0}{0.0} & \msb{50.5}{0.2} & \msb{93.9}{0.1} \\
    \midrule
    4-shot
      & SPADE$^\dagger$ & 84.8 & 91.5 & 92.5 & 92.7 & 46.2 & 87.0 & 81.7 & 82.1 & 83.4 & 96.6 & 43.6 & 87.3 \\
      & PaDiM$^\dagger$ & -- & -- & -- & -- & -- & -- & 72.8 & 78.0 & 75.6 & 93.2 & 24.6 & 72.6 \\
      & PatchCore$^\dagger$ & 88.8 & 92.6 & 94.5 & 94.3 & 55.0 & 84.3 & 85.3 & 84.3 & 87.5 & 96.8 & 43.9 & 84.9 \\
      & FastRecon & 91.3 & -- & -- & 96.1 & -- & -- & 88.2 & -- & -- & \underline{98.0} & -- & -- \\
      & WinCLIP+ & 95.2 & 94.7 & 97.3 & 96.2 & 59.5 & 89.0 & 87.3 & 84.2 & 88.8 & 97.2 & 47.0 & 87.6 \\
      & APRIL-GAN$^\ddagger$ & 92.8 & 92.8 & 96.3 & 95.9 & 56.9 & 91.8 & 92.6 & 88.4 & 94.5 & 96.2 & 40.0 & 90.2 \\
      & ADP$_\ell$ & 96.2 & -- & -- & -- & -- & -- & 88.4 & -- & -- & -- & -- & -- \\
      & SubspaceAD-R$^*$ & 97.3 & \underline{97.2} & \underline{98.6} & \textbf{97.6} & \underline{60.7} & \textbf{94.6} & 91.6 & 88.7 & 90.6 & 97.9 & 40.3 & \underline{94.1} \\
      & AnomalyDINO$^*$ & \underline{97.7} & 97.0 & 98.5 & 96.4 & 58.3 & 92.9 & \underline{94.1} & \underline{89.8} & \underline{95.0} & 97.8 & \underline{47.9} & 93.0 \\
      & \textbf{Ours} & \msb{98.3}{0.1} & \msb{97.5}{0.2} & \msb{99.2}{0.1} & \underline{\ms{97.5}{0.1}} & \msb{61.4}{0.3} & \underline{\ms{94.4}{0.2}} & \msb{94.9}{0.6} & \msb{91.1}{0.4} & \msb{95.7}{0.6} & \msb{98.2}{0.0} & \msb{51.3}{0.2} & \msb{94.3}{0.1} \\
    \midrule
    8-shot
      & WinCLIP+$^\S$ & 94.6 & -- & -- & -- & -- & -- & 85.0 & -- & -- & -- & -- & -- \\
      & ADP$_\ell$ & 97.0 & -- & -- & -- & -- & -- & 89.2 & -- & -- & -- & -- & -- \\
      & SubspaceAD-R$^*$ & 97.7 & 97.2 & 98.8 & \underline{97.6} & \underline{61.3} & \textbf{94.7} & 92.7 & 89.4 & 92.1 & \underline{98.0} & 41.6 & \underline{94.3} \\
      & AnomalyDINO$^*$ & \underline{98.2} & \underline{97.6} & \underline{98.9} & 96.7 & 59.0 & 93.2 & \underline{94.9} & \underline{90.8} & \underline{95.8} & 97.9 & \underline{49.3} & 93.5 \\
      & \textbf{Ours} & \msb{98.7}{0.0} & \msb{98.0}{0.1} & \msb{99.3}{0.0} & \msb{97.7}{0.0} & \msb{61.8}{0.2} & \msb{94.7}{0.0} & \msb{95.5}{0.1} & \msb{92.0}{0.2} & \msb{96.2}{0.1} & \msb{98.3}{0.0} & \msb{52.5}{0.3} & \msb{94.6}{0.1} \\
    \midrule
    16-shot
      & WinCLIP+$^\S$ & 94.8 & -- & -- & -- & -- & -- & 85.0 & -- & -- & -- & -- & -- \\
      & ADP$_\ell$ & 97.0 & -- & -- & -- & -- & -- & 90.1 & -- & -- & -- & -- & -- \\
      & SubspaceAD-R$^*$ & 97.8 & 97.4 & 98.9 & \underline{97.7} & \underline{61.7} & \underline{94.8} & 93.5 & 89.9 & 93.2 & \underline{98.0} & 42.1 & \underline{94.4} \\
      & AnomalyDINO$^*$ & \underline{98.5} & \underline{98.0} & \underline{99.3} & 96.9 & 59.6 & 93.4 & \underline{95.7} & \underline{91.8} & \underline{96.3} & \underline{98.0} & \underline{50.0} & 93.9 \\
      & \textbf{Ours} & \msb{98.9}{0.1} & \msb{98.2}{0.1} & \msb{99.5}{0.1} & \msb{97.8}{0.0} & \msb{62.3}{0.0} & \msb{94.9}{0.1} & \msb{96.1}{0.1} & \msb{92.9}{0.2} & \msb{96.7}{0.1} & \msb{98.3}{0.0} & \msb{52.9}{0.3} & \msb{94.9}{0.0} \\
    \bottomrule
  \end{tabular}}
\end{table*}

\paragraph{MVTec-AD (Table~\ref{tab:main_results}).}
Our method leads across all five shot counts, progressing from $96.4\% / 98.3\%$ I-AUROC / I-AP at $1$-shot to $98.3\% / 99.2\%$ at $4$-shot and $98.9\% / 99.5\%$ at $16$-shot, while achieving $97.2\%$--$97.8\%$ P-AUROC, $59.5\%$--$62.3\%$ P-$F_1$-max, and $93.9\%$--$94.9\%$ AU-PRO. It also outperforms our $3$-seed official SubspaceAD-R$^*$ reproduction on I-AUROC ($+1.0$ to $+1.8$), I-AP ($+0.6$ to $+0.9$), and P-$F_1$-max ($+0.5$ to $+1.6$) across all five shot counts.

\paragraph{VisA (Table~\ref{tab:main_results}).}
Gains are equally consistent under native-resolution evaluation. At $1$-shot, our method lifts the same-backbone AnomalyDINO$^*$ baseline by $+2.01$ I-AUROC ($93.04\%$), $+4.43$ P-$F_1$-max ($48.92\%$), and $+1.74$ AU-PRO ($93.39\%$) while cutting $1$-shot I-AUROC seed std nearly in half ($0.50\% \to 0.27\%$). Performance scales monotonically through $16$-shot ($96.11\%$ I-AUROC, $52.87\%$ P-$F_1$-max, $94.89\%$ AU-PRO), while exceeding official SubspaceAD-R$^*$ by $+2.6$ to $+3.7$ I-AUROC, $+3.5$ to $+5.6$ I-AP, and $+10.8$ to $+11.1$ P-$F_1$-max across $1$--$16$-shot.

\subsection{Ablation Studies}
\label{sec:exp_ablation}

\begin{figure*}[!t]
  \centering
  \includegraphics[width=0.63\textwidth]{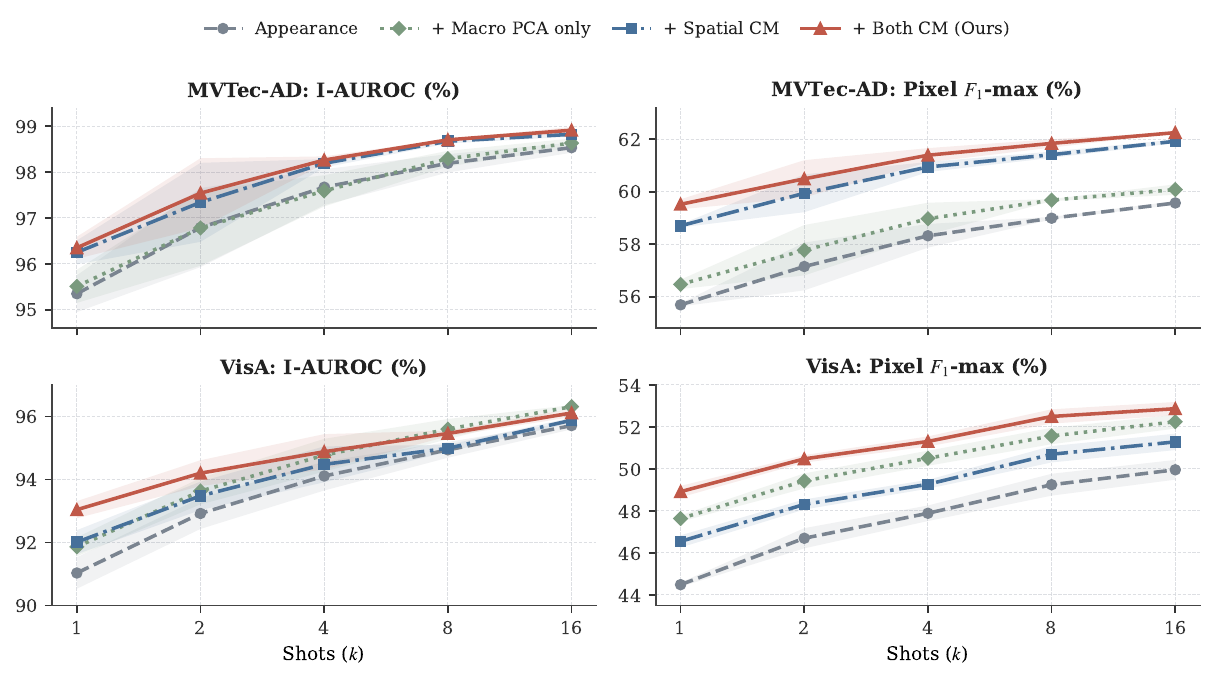}
  \caption{Component scaling curves on \textbf{MVTec-AD} (top) and \textbf{VisA} (bottom) across $k \in \{1, 2, 4, 8, 16\}$ ($3$-seed mean $\pm$ std). Spatial common-mode removal and macro PCA attenuation provide complementary gains across both benchmarks.}
  \label{fig:scaling}
\end{figure*}

\paragraph{Component Ablation (Spatial vs.\ Distributional Common-Mode Suppression).}
Table~\ref{tab:comp_ablation} and Figure~\ref{fig:scaling} isolate all four configurations across both benchmarks under the unified few-shot protocol: the pure appearance baseline (Appearance), macro-only PCA attenuation (+ Macro PCA only), micro-only spatial common-mode removal (+ Spatial CM), and our full dual-scale model (+ Both CM). The two scales exhibit clear complementarity: on MVTec-AD, spatial common-mode removal drives the primary jump ($+0.90$ I-AUROC, $+3.01$ P-$F_1$-max, $+1.94$ AU-PRO at $1$-shot), while adding macro PCA further lifts P-AUROC ($96.91\% \to 97.23\%$) and P-$F_1$-max ($58.70\% \to 59.52\%$); on VisA, macro PCA attenuation provides a strong standalone boost ($+3.15$ P-$F_1$-max at $1$-shot), and combining both scales (+ Both CM) achieves the strongest overall balance across all $1$--$16$-shot regimes ($+0.65$ I-AUROC / $+3.15$ P-$F_1$-max / $+1.64$ AU-PRO on MVTec-AD, and $+1.00$ / $+3.56$ / $+1.33$ on VisA on average over the appearance baseline).

\begin{table*}[!t]
  \centering
  \caption{Component ablation on \textbf{MVTec-AD} and \textbf{VisA} ($3$-seed mean $\pm$ std in $\%$).}
  \label{tab:comp_ablation}
  \footnotesize
  \renewcommand{\arraystretch}{1.00}
  \setlength{\tabcolsep}{11.0pt}
  \resizebox{0.88\textwidth}{!}{%
  \begin{tabular}{llccc@{\hspace{18pt}}ccc@{\hspace{12pt}}}
    \toprule
    & & \multicolumn{3}{c}{\textbf{Image-level}} & \multicolumn{3}{c}{\textbf{Pixel-level}} \\
    \cmidrule(lr){3-5} \cmidrule(lr){6-8}
    \textbf{Dataset / Setup} & \textbf{Variant} & AUROC & $F_1$-max & AP & AUROC & $F_1$-max & PRO \\
    \midrule
    MVTec 1-shot
      & Appearance & \ms{95.35}{0.40} & \ms{95.50}{0.22} & \ms{97.04}{0.20} & \ms{96.02}{0.07} & \ms{55.69}{0.06} & \ms{91.86}{0.20} \\
      & + Macro PCA only & \ms{95.51}{0.36} & \ms{95.80}{0.34} & \ms{97.34}{0.07} & \ms{96.39}{0.07} & \ms{56.47}{0.18} & \ms{92.02}{0.23} \\
      & + Spatial CM & \ms{96.25}{0.26} & \ms{96.54}{0.23} & \ms{98.09}{0.24} & \ms{96.91}{0.07} & \ms{58.70}{0.06} & \ms{93.80}{0.12} \\
      & \textbf{+ Both CM (Ours)} & \msb{96.35}{0.24} & \msb{96.68}{0.29} & \msb{98.32}{0.17} & \msb{97.23}{0.08} & \msb{59.52}{0.19} & \msb{93.87}{0.14} \\
    \midrule
    MVTec 4-shot
      & Appearance & \ms{97.67}{0.39} & \ms{97.01}{0.41} & \ms{98.54}{0.62} & \ms{96.44}{0.12} & \ms{58.32}{0.45} & \ms{92.88}{0.10} \\
      & + Macro PCA only & \ms{97.59}{0.34} & \ms{97.11}{0.50} & \ms{98.58}{0.44} & \ms{96.78}{0.12} & \ms{58.97}{0.61} & \ms{92.96}{0.11} \\
      & + Spatial CM & \ms{98.19}{0.09} & \ms{97.50}{0.28} & \ms{99.08}{0.22} & \ms{97.23}{0.10} & \ms{60.94}{0.19} & \msb{94.50}{0.13} \\
      & \textbf{+ Both CM (Ours)} & \msb{98.26}{0.08} & \msb{97.53}{0.24} & \msb{99.19}{0.11} & \msb{97.49}{0.09} & \msb{61.39}{0.27} & \ms{94.44}{0.15} \\
    \midrule
    MVTec 16-shot
      & Appearance & \ms{98.53}{0.11} & \ms{97.97}{0.06} & \ms{99.25}{0.07} & \ms{96.89}{0.02} & \ms{59.57}{0.07} & \ms{93.44}{0.07} \\
      & + Macro PCA only & \ms{98.63}{0.10} & \ms{98.03}{0.12} & \ms{99.25}{0.14} & \ms{97.17}{0.03} & \ms{60.08}{0.20} & \ms{93.31}{0.07} \\
      & + Spatial CM & \ms{98.82}{0.05} & \ms{98.10}{0.08} & \ms{99.46}{0.04} & \ms{97.60}{0.02} & \ms{61.92}{0.07} & \msb{95.00}{0.03} \\
      & \textbf{+ Both CM (Ours)} & \msb{98.91}{0.07} & \msb{98.21}{0.11} & \msb{99.49}{0.06} & \msb{97.79}{0.02} & \msb{62.25}{0.02} & \ms{94.89}{0.05} \\
    \midrule
    VisA 1-shot
      & Appearance & \ms{91.03}{0.50} & \ms{87.52}{0.59} & \ms{92.45}{0.47} & \ms{97.28}{0.03} & \ms{44.49}{0.10} & \ms{91.65}{0.22} \\
      & + Macro PCA only & \ms{91.86}{0.30} & \ms{88.16}{0.27} & \ms{93.09}{0.47} & \ms{97.63}{0.03} & \ms{47.64}{0.15} & \ms{92.43}{0.24} \\
      & + Spatial CM & \ms{92.01}{0.38} & \ms{88.16}{0.17} & \ms{93.45}{0.28} & \ms{97.64}{0.02} & \ms{46.55}{0.29} & \ms{92.80}{0.18} \\
      & \textbf{+ Both CM (Ours)} & \msb{93.04}{0.27} & \msb{89.41}{0.12} & \msb{94.18}{0.41} & \msb{97.86}{0.02} & \msb{48.92}{0.27} & \msb{93.39}{0.22} \\
    \midrule
    VisA 4-shot
      & Appearance & \ms{94.11}{0.46} & \ms{89.76}{0.41} & \ms{94.99}{0.48} & \ms{97.75}{0.01} & \ms{47.89}{0.37} & \ms{93.01}{0.07} \\
      & + Macro PCA only & \ms{94.77}{0.52} & \ms{90.68}{0.43} & \ms{95.44}{0.55} & \ms{97.97}{0.01} & \ms{50.51}{0.36} & \ms{93.57}{0.12} \\
      & + Spatial CM & \ms{94.48}{0.57} & \ms{90.53}{0.42} & \ms{95.40}{0.64} & \ms{98.02}{0.01} & \ms{49.27}{0.21} & \ms{93.88}{0.09} \\
      & \textbf{+ Both CM (Ours)} & \msb{94.88}{0.56} & \msb{91.10}{0.41} & \msb{95.71}{0.57} & \msb{98.15}{0.01} & \msb{51.31}{0.24} & \msb{94.27}{0.08} \\
    \midrule
    VisA 16-shot
      & Appearance & \ms{95.71}{0.12} & \ms{91.80}{0.33} & \ms{96.31}{0.11} & \ms{98.03}{0.01} & \ms{49.96}{0.47} & \ms{93.90}{0.01} \\
      & + Macro PCA only & \msb{96.31}{0.05} & \ms{92.66}{0.04} & \msb{96.83}{0.10} & \ms{98.15}{0.02} & \ms{52.25}{0.33} & \ms{94.20}{0.03} \\
      & + Spatial CM & \ms{95.88}{0.08} & \ms{92.13}{0.06} & \ms{96.58}{0.12} & \ms{98.26}{0.01} & \ms{51.29}{0.39} & \ms{94.68}{0.00} \\
      & \textbf{+ Both CM (Ours)} & \ms{96.11}{0.13} & \msb{92.85}{0.23} & \ms{96.71}{0.11} & \msb{98.32}{0.01} & \msb{52.87}{0.32} & \msb{94.89}{0.03} \\
    \bottomrule
  \end{tabular}}
\end{table*}

\paragraph{Hyperparameter Sensitivity and Calibration.}
Table~\ref{tab:sensitivity} (Appendix~\ref{app:extra}) evaluates $4$-shot sensitivity to $(k_{\text{pca}}, \alpha, w)$ and $1$/$4$-shot reference-only calibration. Across both benchmarks, performance remains highly stable around the single global configuration $(k_{\text{pca}}{=}64, \alpha{=}0.5, w{=}0.25)$, forming a broad plateau across $k_{\text{pca}} \in [32, 128]$, $\alpha \in [0.25, 0.75]$, and $w \in [0.15, 0.50]$. Notably, hard removal ($\alpha{=}1.0$) degrades AU-PRO on both benchmarks ($94.44\% \to 93.58\%$ on MVTec-AD; $94.27\% \to 93.05\%$ on VisA), confirming the necessity of soft attenuation ($0 < \alpha < 1$). Reference-only quantile calibration consistently improves pixel-level P-AUROC and AU-PRO across both datasets and shot counts such as $+0.12$ P-AUROC and $+0.15$--$0.16$ AU-PRO on VisA as well as all MVTec-AD image-level metrics, establishing a well-balanced default pipeline.

\paragraph{Runtime and Efficiency (Table~\ref{tab:efficiency}).}
Table~\ref{tab:efficiency} benchmarks end-to-end offline bank construction and per-image inference latency against the pure appearance baseline. Consistent with our $O(N)$ per-image ($O(1)$ per-patch) extra scoring complexity, dual-scale common-mode suppression adds only $+5.18$\,ms/img ($+7.7\%$) on MVTec-AD and $+8.48$\,ms/img ($+8.1\%$) on VisA, while one-time offline bank construction completes in $5.04$--$7.37$\,s per category.

\begin{table}[H]
  \centering
  \caption{Runtime comparison on MVTec-AD and VisA: offline bank construction (s) and per-image latency breakdown (ms).}
  \label{tab:efficiency}
  \scriptsize
  \renewcommand{\arraystretch}{1.14}
  \setlength{\tabcolsep}{4.2pt}
  \resizebox{\linewidth}{!}{%
  \begin{tabular}{llcccc}
    \toprule
    \textbf{Dataset} & \textbf{Method} & \textbf{Bank (s)} & \textbf{Latency (ms)} & \textbf{Feat. (ms)} & \textbf{Score (ms)} \\
    \midrule
    MVTec-AD & Appearance & 2.38 & 67.17 & 54.62 & 12.55 \\
             & \textbf{Ours} & 5.04 & 72.35 & 56.46 & 15.89 \\
    \midrule
    VisA     & Appearance & 3.33 & 104.35 & 93.21 & 11.14 \\
             & \textbf{Ours} & 7.37 & 112.83 & 99.10 & 13.73 \\
    \bottomrule
  \end{tabular}}
\end{table}

\paragraph{Qualitative Localization (Appendix~\ref{app:qual}).}
Figure~\ref{fig:qualitative} (Appendix~\ref{app:qual}) presents $4$-shot anomaly localization maps across $16$ representative categories from MVTec-AD ($8$ categories) and VisA ($8$ categories). Across both fine surface/structural defects (e.g., \emph{grid}, \emph{tile}, \emph{wood}, \emph{zipper}) and subtle multi-instance or bent-pin anomalies (e.g., \emph{chewinggum}, \emph{fryum}, \emph{pcb1}), dual-scale common-mode suppression concentrates high anomaly responses tightly within the ground-truth masks.

\section{Conclusion and Limitations}
\label{sec:conclusion}

We introduced dual-scale common-mode suppression, a training-free framework that unifies micro-scale spatial context removal and macro-scale soft distributional PCA attenuation ($I - \alpha P$) within a single frozen DINOv2 forward pass. By attenuating shared neighborhood-average and bank-level anisotropic variance prior to unit-sphere re-normalization, the proposed operator expands discriminative angular margins with less than $8.1\%$ inference latency overhead, yielding consistent detection and localization gains across MVTec-AD and VisA from $1$-shot to $16$-shot under one global hyperparameter setting.

\paragraph{Limitations and Future Work.}
As a training-free framework built on local patch-level retrieval, our method shares two inherent boundaries of this paradigm. First, anomalies defined purely by global logical or compositional rules such as missing parts, wrong object counts, or long-range structural violations where every individual patch remains locally normal, remain difficult to capture through local patch neighborhoods alone; pairing training-free hyperspherical suppression with global object-centric or structural reasoning is a natural next step. Second, because inference relies on frozen visual features and fixed reference views, severe out-of-plane 3D viewpoint shifts or subtle functional defects not salient in standard RGB imagery remain challenging, motivating future integration with multi-view geometric or spectrum-adaptive conditioning.

\clearpage
\twocolumn[{%
\section*{Impact Statement}
This paper presents work whose goal is to advance training-free industrial visual anomaly detection and localization. Automated defect inspection improves manufacturing quality control and reduces material waste; we do not foresee negative societal consequences beyond standard considerations for industrial computer vision systems.
\vspace{10pt}
}]

\appendix
\newpage
\section{Symbol Reference}
\label{app:symbols}
\begin{table}[H]
  \centering
  \caption{Summary of notation used in Sections~\ref{sec:prelim}--\ref{sec:method}.}
  \footnotesize
  \renewcommand{\arraystretch}{1.04}
  \setlength{\tabcolsep}{6pt}
  \begin{tabularx}{\linewidth}{l>{\raggedright\arraybackslash}X}
    \toprule
    \textbf{Symbol} & \textbf{Meaning} \\
    \midrule
    $k,\, V$ & Reference shots ($\{1,2,4,8,16\}$) and augmented views \\
    $N,\, D$ & Patches per image ($H_p W_p$, $1024$ on square inputs) and dim ($D=768$) \\
    $I$ & $D\times D$ identity matrix \\
    $f_i,\, y_i^{(b)}$ & Last-block appearance and block-$b$ normalized feature \\
    $A_{ij}^{(b)},\, c_i^{(b)},\, r_i^{(b)}$ & Attention weight, unit context vector, orthogonal residual \\
    $\mathcal{B}=\{7,8,9\}$ & Relation transformer blocks ($0$-indexed) \\
    $a(q),\, \rel_b(q),\, e_b(q)$ & Appearance score, matched relation distance, calibrated excess \\
    $\tau_b,\, \sigma_b,\, s_{\text{app}}$ & Reference-only calibration threshold, spread, appearance scale \\
    $w$ & Relation fusion weight ($w=0.25$) \\
    $\mu,\, \Sigma,\, U$ & Reference mean, covariance, top-$k_{\text{pca}}$ eigenvectors \\
    $k_{\text{pca}},\, \alpha$ & Suppressed directions ($64$) and strength ($0.5$) \\
    $M,\, T_{1\%}(\cdot)$ & Test foreground mask and top-$1\%$ mean operator \\
    \bottomrule
  \end{tabularx}
\end{table}

\clearpage
\onecolumn
\section{Hyperparameter Sensitivity, Calibration, and Backbone Scaling}
\label{app:extra}

Table~\ref{tab:sensitivity} details $4$-shot hyperparameter sensitivity to $(k_{\text{pca}}, \alpha, w)$, $1$/$4$-shot reference-only calibration, and $4$-shot DINOv2 backbone scaling across ViT-S/14 ($D{=}384$), ViT-B/14 ($D{=}768$, default), and ViT-L/14 ($D{=}1024$, extracting relational features at proportional mid-layer blocks $14,16,18$ with $w{=}0.15$ on MVTec-AD and $w{=}0.10$ on VisA). The lightweight ViT-S/14 matches or surpasses ViT-B/14 on pixel localization ($62.43\%$ P-$F_1$ on MVTec-AD, $98.38\%$ P-AUROC on VisA), while ViT-L/14 further lifts VisA I-AUROC to $95.88\%$ and MVTec-AD AU-PRO to $95.56\%$.

\begin{table}[H]
  \centering
  \caption{Hyperparameter sensitivity ($4$-shot), calibration ablation ($1$/$4$-shot), and DINOv2 backbone scaling ($4$-shot) on \textbf{MVTec-AD} and \textbf{VisA} (in $\%$; hyperparameter/calibration blocks report $3$-seed mean $\pm$ std). Default global setting is shaded; best per block in \textbf{bold}.}
  \label{tab:sensitivity}
  \scriptsize
  \renewcommand{\ms}[2]{#1{\tiny\color{black!60}\,$\pm$\,#2}}
  \renewcommand{\msb}[2]{\textbf{#1}{\tiny\color{black!60}\,$\pm$\,#2}}
  \renewcommand{\arraystretch}{0.96}
  \setlength{\tabcolsep}{8.0pt}
  \resizebox{\textwidth}{!}{%
  \begin{tabular}{llcccccc@{\hspace{22pt}}cccccc}
    \toprule
    & & \multicolumn{6}{c}{\textbf{MVTec-AD} ($15$ categories)} & \multicolumn{6}{c}{\textbf{VisA} ($12$ categories)} \\
    \cmidrule(lr){3-8} \cmidrule(lr){9-14}
    \textbf{Study} & \textbf{Setting} & I-AUC & I-$F_1$ & I-AP & P-AUC & P-$F_1$ & PRO & I-AUC & I-$F_1$ & I-AP & P-AUC & P-$F_1$ & PRO \\
    \midrule
    $k_{\text{pca}}$ ($4$-shot)
      & $k{=}16, \alpha{=}0.5$ & \ms{98.16}{0.17} & \ms{97.45}{0.31} & \ms{99.17}{0.15} & \ms{97.29}{0.10} & \ms{60.63}{0.32} & \ms{94.23}{0.17} & \ms{94.15}{0.60} & \ms{90.52}{0.41} & \ms{95.19}{0.65} & \ms{98.06}{0.02} & \ms{50.05}{0.17} & \ms{93.61}{0.16} \\
      & $k{=}32, \alpha{=}0.5$ & \ms{98.25}{0.10} & \msb{97.64}{0.17} & \msb{99.24}{0.10} & \ms{97.41}{0.10} & \ms{61.03}{0.29} & \ms{94.34}{0.17} & \ms{94.41}{0.47} & \ms{90.67}{0.28} & \ms{95.27}{0.51} & \ms{98.11}{0.01} & \ms{50.71}{0.23} & \ms{93.84}{0.13} \\
      \rowcolor{gray!12} & $k{=}64, \alpha{=}0.5$ & \msb{98.26}{0.08} & \ms{97.53}{0.24} & \ms{99.19}{0.11} & \msb{97.49}{0.09} & \ms{61.39}{0.27} & \ms{94.44}{0.15} & \ms{94.88}{0.56} & \ms{91.10}{0.41} & \ms{95.71}{0.57} & \ms{98.15}{0.01} & \ms{51.31}{0.24} & \ms{94.27}{0.08} \\
      & $k{=}128, \alpha{=}0.5$ & \ms{98.22}{0.10} & \ms{97.48}{0.28} & \ms{99.12}{0.18} & \msb{97.49}{0.10} & \msb{61.48}{0.30} & \msb{94.48}{0.15} & \msb{95.26}{0.50} & \msb{91.37}{0.26} & \msb{96.00}{0.51} & \msb{98.18}{0.00} & \msb{51.51}{0.16} & \msb{94.57}{0.06} \\
    \midrule
    $\alpha$ ($4$-shot)
      & $k{=}64, \alpha{=}0.25$ & \ms{98.19}{0.06} & \ms{97.47}{0.27} & \ms{99.10}{0.18} & \ms{97.37}{0.10} & \ms{61.23}{0.20} & \msb{94.54}{0.13} & \ms{94.75}{0.55} & \ms{90.77}{0.45} & \ms{95.60}{0.61} & \ms{98.11}{0.01} & \ms{50.24}{0.25} & \ms{94.15}{0.08} \\
      \rowcolor{gray!12} & $k{=}64, \alpha{=}0.50$ & \msb{98.26}{0.08} & \ms{97.53}{0.24} & \msb{99.19}{0.11} & \msb{97.49}{0.09} & \msb{61.39}{0.27} & \ms{94.44}{0.15} & \msb{94.88}{0.56} & \msb{91.10}{0.41} & \msb{95.71}{0.57} & \msb{98.15}{0.01} & \ms{51.31}{0.24} & \msb{94.27}{0.08} \\
      & $k{=}64, \alpha{=}0.75$ & \ms{98.18}{0.03} & \msb{97.68}{0.23} & \ms{99.14}{0.06} & \ms{97.48}{0.10} & \ms{61.23}{0.42} & \ms{94.06}{0.17} & \ms{94.64}{0.51} & \ms{91.02}{0.31} & \ms{95.33}{0.46} & \ms{98.10}{0.01} & \msb{52.16}{0.26} & \ms{94.01}{0.16} \\
      & $k{=}64, \alpha{=}1.00$ & \ms{98.10}{0.07} & \ms{97.59}{0.23} & \ms{99.09}{0.06} & \ms{97.33}{0.12} & \ms{60.58}{0.63} & \ms{93.58}{0.20} & \ms{93.99}{0.50} & \ms{90.49}{0.54} & \ms{94.72}{0.40} & \ms{97.91}{0.02} & \ms{51.96}{0.32} & \ms{93.05}{0.32} \\
    \midrule
    $w$ ($4$-shot)
      & $w{=}0.00$ & \ms{97.59}{0.34} & \ms{97.11}{0.50} & \ms{98.58}{0.44} & \ms{96.78}{0.12} & \ms{58.97}{0.61} & \ms{92.96}{0.11} & \ms{94.77}{0.52} & \ms{90.68}{0.43} & \ms{95.44}{0.55} & \ms{97.97}{0.01} & \ms{50.51}{0.36} & \ms{93.57}{0.12} \\
      & $w{=}0.15$ & \ms{98.16}{0.08} & \ms{97.49}{0.30} & \ms{99.07}{0.17} & \ms{97.38}{0.11} & \ms{61.12}{0.30} & \ms{94.21}{0.14} & \msb{95.07}{0.55} & \msb{91.16}{0.35} & \msb{95.81}{0.56} & \ms{98.14}{0.01} & \msb{51.41}{0.26} & \ms{94.22}{0.08} \\
      \rowcolor{gray!12} & $w{=}0.25$ & \ms{98.26}{0.08} & \ms{97.53}{0.24} & \ms{99.19}{0.11} & \ms{97.49}{0.09} & \ms{61.39}{0.27} & \ms{94.44}{0.15} & \ms{94.88}{0.56} & \ms{91.10}{0.41} & \ms{95.71}{0.57} & \ms{98.15}{0.01} & \ms{51.31}{0.24} & \msb{94.27}{0.08} \\
      & $w{=}0.35$ & \msb{98.28}{0.07} & \msb{97.59}{0.19} & \ms{99.23}{0.08} & \ms{97.53}{0.08} & \msb{61.40}{0.28} & \ms{94.55}{0.15} & \ms{94.66}{0.54} & \ms{90.84}{0.41} & \ms{95.56}{0.59} & \msb{98.16}{0.01} & \ms{51.08}{0.23} & \ms{94.25}{0.08} \\
      & $w{=}0.50$ & \ms{98.27}{0.09} & \ms{97.49}{0.22} & \msb{99.25}{0.06} & \msb{97.55}{0.07} & \ms{61.27}{0.30} & \msb{94.60}{0.15} & \ms{94.37}{0.51} & \ms{90.47}{0.38} & \ms{95.35}{0.58} & \ms{98.15}{0.01} & \ms{50.73}{0.20} & \ms{94.20}{0.09} \\
    \midrule
    Calib.\ ($1$-shot)
      & Uncalibrated & \ms{96.21}{0.21} & \ms{96.60}{0.35} & \ms{98.07}{0.08} & \ms{97.20}{0.08} & \msb{59.54}{0.19} & \ms{93.73}{0.14} & \msb{93.05}{0.35} & \ms{89.19}{0.09} & \msb{94.21}{0.51} & \ms{97.74}{0.02} & \msb{49.23}{0.23} & \ms{93.24}{0.24} \\
      \rowcolor{gray!12} & Calibrated & \msb{96.35}{0.24} & \msb{96.68}{0.29} & \msb{98.32}{0.17} & \msb{97.23}{0.08} & \ms{59.52}{0.19} & \msb{93.87}{0.14} & \ms{93.04}{0.27} & \msb{89.41}{0.12} & \ms{94.18}{0.41} & \msb{97.86}{0.02} & \ms{48.92}{0.27} & \msb{93.39}{0.22} \\
    \midrule
    Calib.\ ($4$-shot)
      & Uncalibrated & \ms{98.18}{0.10} & \ms{97.48}{0.30} & \ms{99.08}{0.19} & \ms{97.47}{0.09} & \msb{61.42}{0.29} & \ms{94.39}{0.15} & \msb{95.07}{0.55} & \ms{91.09}{0.36} & \msb{95.84}{0.55} & \ms{98.03}{0.01} & \msb{51.70}{0.21} & \ms{94.11}{0.06} \\
      \rowcolor{gray!12} & Calibrated & \msb{98.26}{0.08} & \msb{97.53}{0.24} & \msb{99.19}{0.11} & \msb{97.49}{0.09} & \ms{61.39}{0.27} & \msb{94.44}{0.15} & \ms{94.88}{0.56} & \msb{91.10}{0.41} & \ms{95.71}{0.57} & \msb{98.15}{0.01} & \ms{51.31}{0.24} & \msb{94.27}{0.08} \\
    \midrule
    Backbone ($4$-shot)
      & ViT-S/14 ($D{=}384$) & 98.19 & \textbf{97.73} & 99.15 & \textbf{97.85} & \textbf{62.43} & 94.16 & 93.95 & 90.24 & 94.21 & \textbf{98.38} & 50.70 & 94.52 \\
      \rowcolor{gray!12} & ViT-B/14 ($D{=}768$) & 98.26 & 97.53 & \textbf{99.19} & 97.49 & 61.39 & 94.44 & 94.88 & 91.10 & 95.71 & 98.15 & \textbf{51.31} & 94.27 \\
      & ViT-L/14 ($D{=}1024$) & \textbf{98.31} & 97.52 & \textbf{99.19} & \textbf{97.85} & 62.33 & \textbf{95.56} & \textbf{95.88} & \textbf{92.09} & \textbf{96.55} & 97.51 & 50.80 & \textbf{94.81} \\
    \bottomrule
  \end{tabular}}
\end{table}

\vspace{-4pt}
\section{Qualitative Localization Gallery}
\label{app:qual}

\begin{figure}[H]
  \centering
  \includegraphics[width=0.58\textwidth]{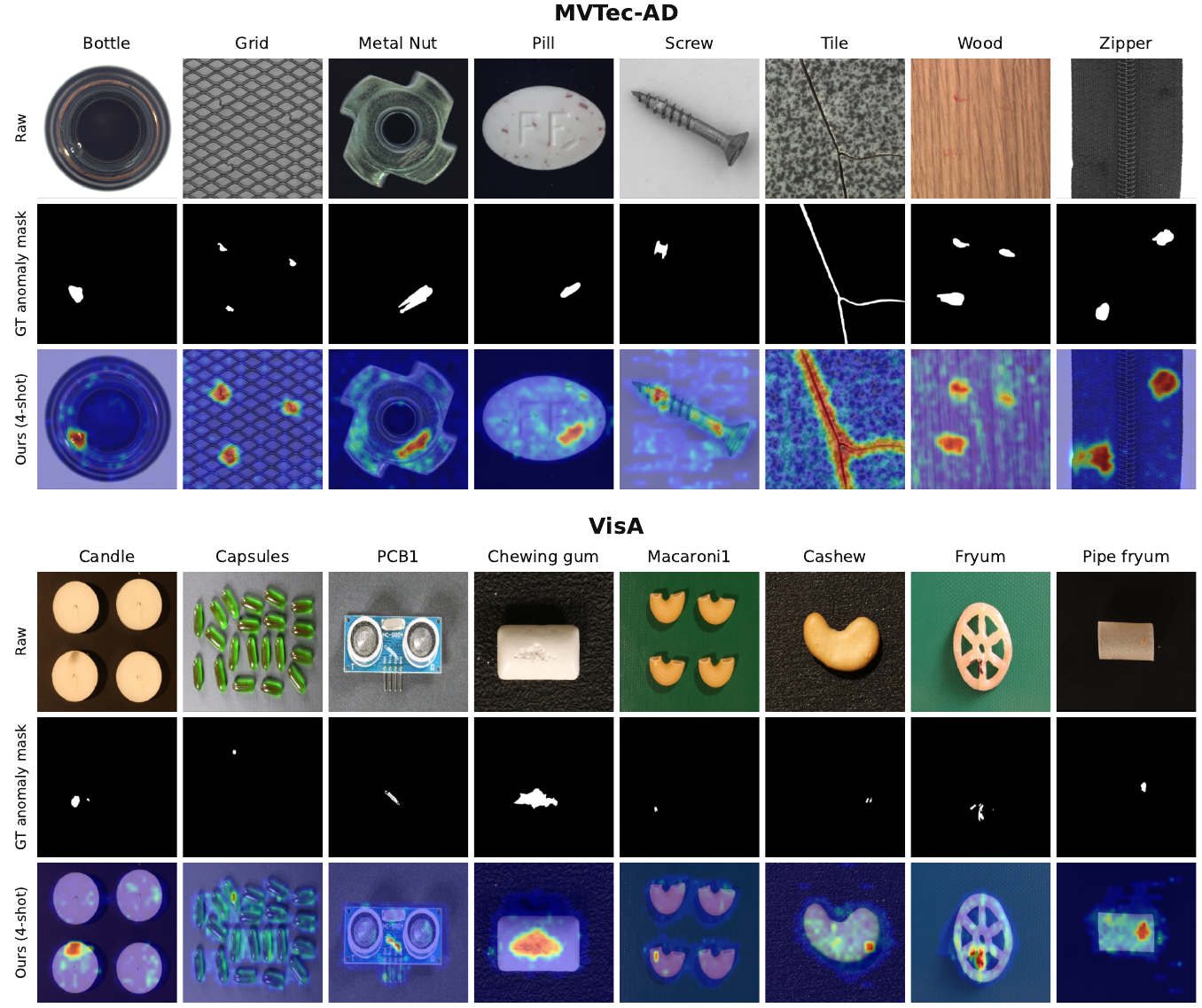}
  \caption{Dual-scale common-mode suppression localizes subtle surface and structural defects tightly within ground-truth masks. Four-shot qualitative results across $16$ categories from MVTec-AD ($8$) and VisA ($8$), showing \textbf{Raw} query, \textbf{GT mask}, and \textbf{Ours}.}
  \label{fig:qualitative}
\end{figure}

Figure~\ref{fig:qualitative} visualizes $4$-shot anomaly localization maps across $16$ categories from MVTec-AD ($8$ categories) and VisA ($8$ categories) under the unified global configuration ($k_{\text{pca}}{=}64, \alpha{=}0.5, w{=}0.25$).

\end{document}